%% file: main.tex
\documentclass{article}
\input{preamble}

\title{ 
    Chronicles-OCR: A Cross-Temporal Perception Benchmark for
    \\
    the Evolutionary Trajectory of Chinese Characters
}

\input{section/author}

\begin{document}
\maketitle

\vspace{-1.0em}

\input{section/author_footnote}
\input{figure_code/teaser}
\input{section/0_abstract}
\input{section/1_introduction}
\input{section/2_related_work}
\input{section/3_benchmark}
\input{section/4_experiments}
\input{section/5_conclusion}

\clearpage
{
    \small
    \setlength{\bibsep}{6pt}
    \bibliographystyle{unsrtnat}
    \bibliography{main}
}

\end{document}

%% file: preamble.tex
\PassOptionsToPackage{numbers,sort&compress}{natbib}
\usepackage{iclr2025_conference,times}
\usepackage{wrapfig}
\usepackage{lipsum} 
\usepackage{wrapfig}

\input{math_commands}

\usepackage{booktabs}      
\usepackage{multirow}      
\usepackage{amssymb}       
\usepackage{graphicx, subfigure}     
\usepackage{colortbl}
\usepackage{xcolor}
\usepackage{hyperref}
\usepackage{url}
\usepackage{multicol}
\usepackage{aliascnt}
\usepackage{adjustbox}
\usepackage{fontawesome5}
\usepackage{CJKutf8}
\usepackage{siunitx} 
\usepackage{enumitem}
\usepackage{amsmath}  
\usepackage{longtable} 
\usepackage{subcaption}
\usepackage{calc}
\usepackage[most]{tcolorbox}
\usepackage{listings}
\usepackage{latexsym}
\usepackage{dsfont}
\usepackage{mathtools}
\usepackage{makecell}
\usepackage[ruled]{algorithm2e}
\usepackage{pifont}

\definecolor{uclablue}{rgb}{0.15, 0.45, 0.68}
\hypersetup{
    breaklinks,
    citecolor=uclablue,
    colorlinks=true,
}

\definecolor{lightgreen}{RGB}{0,150,0}
\definecolor{myred}{RGB}{200,0,0} 
\newcommand{\mypara}[1]{\vspace{-0.2mm}\noindent\textbf{#1}\hspace{0.02cm}}

\usepackage{tabularx}
\robustify\bfseries
\usepackage{subcaption}
\usepackage{cleveref}

\tcbuselibrary{listings,skins,breakable}

\newtcolorbox{prompt}[1][]{enhanced,
    breakable,
    colback=white,              
    colframe=black,             
    coltitle=black,             
    colbacktitle=gray!20,       
    fonttitle=\bfseries,
    title=Prompt,
    #1
}

\newtcolorbox{remark}[1][]{enhanced,
    breakable,
    colback=violet!6,           
    colframe=violet!50!black,   
    coltitle=black,             
    colbacktitle=violet!18,     
    fonttitle=\bfseries,
    title=Remark,
    #1
}

\makeatletter

\let\chron@oldtitle\title
\renewcommand{\title}[1]{\chron@oldtitle{#1}\gdef\thetitle{#1}}
\providecommand{\thetitle}{}
\makeatother

\definecolor{linkColor}{rgb}{0.2,0.4,0.6}
\definecolor{myblue}{HTML}{0379AC}
\definecolor{myred}{HTML}{A50E50}
\definecolor{myorange}{RGB}{238, 133, 74}
\definecolor{latentcolor}{named}{cyan}
\definecolor{normalcolor}{RGB}{0, 0, 0}
\usepackage{marvosym}

\newcommand{\cmark}{\ding{51}} 
\newcommand{\nextline}{\\} 
\newcommand{\ul}[1]{\underline{#1}}

\newcommand{\myeg}{\textit{e.g.,}\xspace}
\newcommand{\logo}[1]{%
    \smash{%
        \raisebox{-\dp\strutbox}{%
            \includegraphics[height=\dimexpr\ht\strutbox+\dp\strutbox\relax]{figures/logo/org/#1}%
        }%
    }\,%
}

\definecolor{table_ours}{HTML}{F5FFFA}   
\definecolor{titleblue}{HTML}{5AA1BF}     
\definecolor{lightblue}{HTML}{EAF4FB}     

\AtEndPreamble{
    \crefname{section}{Sec.}{Secs.}         \Crefname{section}{Sec.}{Secs.}
    \crefname{equation}{Eq.}{Eqs.}          \Crefname{equation}{Eq.}{Eqs.}
    \crefname{table}{Tab.}{Tabs.}           \Crefname{table}{Tab.}{Tabs.}
    \crefname{figure}{Fig.}{Figs.}          \Crefname{figure}{Fig.}{Figs.}
    \crefname{promptcount}{prompt}{prompts} \Crefname{promptcount}{Prompt}{Prompts}
    \creflabelformat{equation}{#2(#1)#3}
    \creflabelformat{table}{#2#1#3}
    \creflabelformat{figure}{#2#1#3}
}

%% file: math_commands.tex
\usepackage{amsmath,amsfonts,bm}
\usepackage{multicol}

\def\eqref#1{equation~\ref{#1}}

\def\1{\bm{1}}

\DeclareMathAlphabet{\mathsfit}{\encodingdefault}{\sfdefault}{m}{sl}
\SetMathAlphabet{\mathsfit}{bold}{\encodingdefault}{\sfdefault}{bx}{n}



%% file: section/author.tex
\author{
    Gengluo Li$^{1}$\quad
    Shangpin Peng$^{2}$\quad
    Xingyu Wan$^{2}$\quad
    Chengquan Zhang$^{2,\, \ddagger}$\quad
    Hao Feng$^{2}$\quad
    Xin Xu$^{2}$
    \\[2pt]
    Pian Wu$^{2}$\quad
    Bang Li$^{3}$\quad
    Zengmao Ding$^{3}$\quad
    Yongge Liu$^{3}$\quad
    Yipei Ye$^{4}$ \quad Yang Yang$^{4}$
    \\[2pt]
    Zhan Shu$^{2}$\quad
    Guojun Yan$^{2}$\quad
    Zhe Li$^{2}$\quad
    Can Ma$^{1}$\quad
    Weiping Wang$^{1}$\quad
    Yu Zhou$^{5,\,}$\textsuperscript{\scalebox{0.8}{\faEnvelope}}\quad
    Han Hu$^{2,\,}$\textsuperscript{\scalebox{0.8}{\faEnvelope}}
    \\[2pt]
    \textbf{$^1$Institute of Information Engineering, Chinese Academy of Sciences}
    \\
    \textbf{$^2$LLM Department, Tencent}
    \quad
    \textbf{$^3$Anyang Normal University}
    \\
    \textbf{$^4$The Palace Museum}
    \quad
    \textbf{$^5$ Nankai University}
    \\[2pt]
    \centerline {
        \tt\small
        ligengluo@iie.ac.cn \quad yzhou@nankai.edu.cn
    }
}

%% file: section/author_footnote.tex
\let\oldthefootnote\thefootnote
\let\thefootnote\relax\footnotetext{
    $^{\ddagger}$ Project leader
    \qquad
    \textsuperscript{\scalebox{0.9}{\faEnvelope}} Corresponding author
}
\let\thefootnote\oldthefootnote

%% file: figure_code/teaser.tex
\begin{figure*}[ht]
    \centering
    \includegraphics[width=\linewidth]{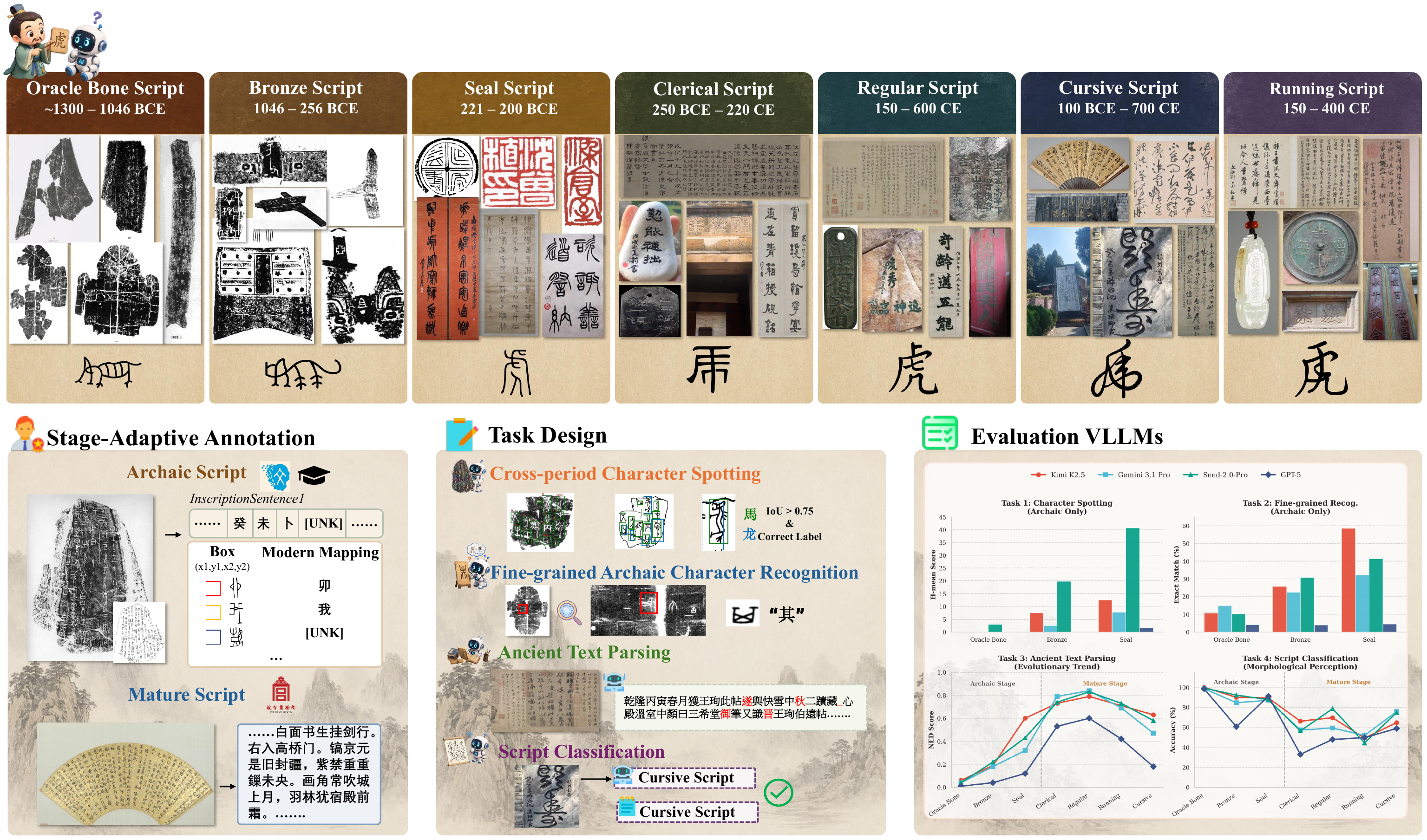}
    \vspace{-1.8em}
    \captionsetup{font={small}}
    \caption{\textbf{Chronicles-OCR overview.}
        The top row showcases diverse physical artifact samples from Chronicles-OCR across seven script stages, alongside the morphological evolution of the modern Chinese character ``\begin{CJK*}{UTF8}{gbsn}虎\end{CJK*}'' (Tiger).
        To comprehensively evaluate VLLMs, we introduce a stage-adaptive annotation paradigm and four progressive tasks. Evaluation results reveal substantial capability gaps in the fine-grained visual perception of archaic scripts.
    }
    \label{fig:teaser}
\end{figure*}

%% file: section/0_abstract.tex
\begin{abstract}
    Vision Large Language Models (VLLMs) have achieved remarkable success in modern text-rich visual understanding. However, their perceptual robustness in the face of the continuous morphological evolution of historical writing systems remains largely unexplored. Existing ancient text datasets typically focus on isolated historical periods, failing to capture the systematic visual distribution shifts spanning thousands of years.
    To bridge this gap and empower Digital Humanities, we introduce \textbf{Chronicles-OCR}, the first comprehensive benchmark specifically designed to evaluate the cross-temporal visual perception capabilities of VLLMs across the complete evolutionary trajectory of Chinese characters, known as the ``Seven Chinese Scripts''. Curated in collaboration with top-tier institutional domain experts, the dataset comprises 2,800 strictly balanced images encompassing highly diverse physical media, ranging from tortoise shells to paper-based calligraphy.
    To accommodate the drastic morphological and topological variations across different historical stages, we propose a novel \textit{Stage-Adaptive Annotation Paradigm}. Based on this, Chronicles-OCR formulates four rigorous quantitative tasks: cross-period character spotting, fine-grained archaic character recognition via visual referring, ancient text parsing, and script classification. By isolating visual perception from semantic reasoning, Chronicles-OCR provides an authoritative platform to expose the limitations of current VLLMs, paving the way for robust, evolution-aware historical text perception. Chronicles-OCR is publicly available at \url{https://github.com/VirtualLUOUCAS/Chronicles-OCR}.
\end{abstract}

%% file: section/1_introduction.tex
\section{Introduction}
\label{sec:introduction}

In recent years, Vision Large Language Models (VLLMs) have achieved remarkable success in modern text-rich visual understanding~\cite{Qwen3_VL_2025, Qwen2_5_VL_2025}.
When evaluated on comprehensive modern benchmarks, such as OCRBench~\cite{OCRBench_v2_2024}, OmniDocBench~\cite{Omnidocbench_2025}, and OCR-Reasoning~\cite{OCR_Reasoning_2025} that span from basic text perception to advanced visual reasoning, state-of-the-art models consistently exhibit exceptional performance~\cite{PaddleOCR_VL_2025, HunyuanOCR_2025}. This success highlights the mature capabilities of current VLLMs in processing and comprehending modern standardized documents. However, this remarkable proficiency heavily relies on the structural consistency of modern writing systems. In contemporary typography and standardized writing, Chinese characters possess unified topological structures and well-defined boundaries, and are predominantly rendered on clean digital media~\cite{MCS_Bench_2025}. Such an ideal setting allows models to learn robust and highly generalizable feature representations through massive pre-training on modern datasets.

Tracing back through history, these structural prerequisites progressively dissolve, revealing a profound transformation in the Chinese character morphology~~\cite{EVOBC_2024, Graph_evolution_2025, Evolution_fewshot_2022}. From the Regular script back to the Running and Cursive scripts, characters begin to exhibit extensive cursive strokes and radical simplifications. Progressing further back to the Clerical and Seal scripts, their topological structures diverge fundamentally from modern conventions. In their most archaic forms, such as the Oracle Bone and Bronze scripts, texts manifest as unstandardized, intricate carved symbols characterized by extreme morphological variance and unconstrained spatial layouts~\cite{OBI_Bench_2025, Oracle_Survey_2026, OBC306_2019, survey_oracle_2024}. Furthermore, these historical texts are inextricably embedded within highly diverse and uncontrolled physical media, ranging from wooden slips and ancient paintings to bronze artifacts, stone steles, and rubbings, which introduces severe background noise and material degradation.
Ultimately, as we traverse backward in time, the text paradigm devolves from a standardized symbolic system into highly unstructured, variable visual representations entangled in complex, degraded physical scenes~\cite{Historical_document_2020}.

This profound distribution shift signifies that the challenge confronting VLLMs extends far beyond processing standard modern documents, requiring precise character spotting, fine-grained archaic recognition, and ancient text parsing amidst unconstrained layouts and drastically varying morphologies.
This requires highly robust fine-grained visual feature extraction and cross-domain generalization capabilities to navigate millennia of morphological changes~\cite{TongGu_VL_2025, OracleAgent_2025, V_oracle_2025, WenyanGPT_2025}. Although previous studies have established datasets for discrete historical slices, such as Oracle Bone script recognition~\cite{OBI_Bench_2025, HWOBC_2020, OracleFusion_2025, HUST_OBC_2024} or Ming and Qing dynasty document OCR~\cite{CASIA_AHCDB_2019}, these efforts remain confined to isolated historical periods, failing to capture the systematic visual evolution. To date, the community lacks a holistic evaluation benchmark spanning the complete evolutionary trajectory of Chinese scripts. When confronted with this extreme historical distribution shift, where do the true perceptual boundaries of current VLLMs lie? What are their foundational perceptual bottlenecks as they traverse backward through time? A unified, quantitative evaluation platform to answer these critical questions remains conspicuously absent.

To bridge this evaluation gap and drive the application of VLLMs in Digital Humanities, we propose \textbf{Chronicles-OCR} (illustrated in~\cref{fig:teaser}), the first comprehensive evaluation benchmark covering the full lifecycle of Chinese script evolution. Systematically encompassing the ``Seven Chinese Scripts'' (Oracle Bone, Bronze, Seal, Clerical, Regular, Running, and Cursive scripts), this benchmark aims to holistically assess the perceptual robustness of VLLMs against cross-temporal visual distribution shifts. Tailored to accommodate the drastic layout and morphological variations across different historical periods, we introduce an innovative \textit{Stage-Adaptive Annotation Paradigm} and corresponding evaluation tasks. Specifically, for archaic scripts,
we deploy a fine-grained strategy comprising single-character bounding boxes, visual referring mechanisms, and modern character mappings to support end-to-end spotting and recognition.
For more mature pre-modern scripts, the evaluation naturally shifts to sequence-level layout comprehension through ancient text parsing. Furthermore, we introduce a universal script classification task across all seven scripts to probe the models' macro-level understanding of morphological evolution. Leveraging this benchmark, we conduct an extensive and in-depth evaluation of mainstream open-source and closed-source VLLMs.

In summary, the primary contributions of this paper are threefold:
\begin{itemize}[
        label=\raisebox{0.5ex}{\tiny$\bullet$},
        leftmargin=1em,
        itemsep=2pt, 
        parsep=0pt, 
        topsep=0pt, 
        partopsep=0pt 
    ]
    \item \textbf{Chronicles-OCR, the First Benchmark Covering the ``Seven Chinese Scripts'':} We introduce Chronicles-OCR to bridge the evaluation gap in historical text perception, achieving the first full-timespan coverage from unstandardized archaic symbols to mature pre-modern scripts.
    \item \textbf{Stage-Adaptive Evaluation Tasks:} Addressing the morphological evolution across different scripts, we innovatively design four evaluation tasks. This strategy transitions from cross-period character spotting and fine-grained visual referring recognition for archaic scripts to complex ancient text parsing for mature scripts, complemented by a universal script classification task. This rigorously reflects the unique perceptual challenges of each historical stage and practical Digital Humanities scenarios.
    \item \textbf{Comprehensive Benchmarking of Perceptual Deficiencies across Eras:} Through extensive evaluations of mainstream VLLMs, we systematically establish performance baselines and quantify their perceptual bottlenecks when confronting historically evolved texts. By revealing the precise limitations of current models across different historical stages, we provide clear optimization trajectories for capacity building in the Digital Humanities.
\end{itemize}

%% file: section/2_related_work.tex
\section{Related Work}
\label{sec:related_work}

\subsection{Evolution and Visual Characteristics of Chinese Scripts}

The morphological evolution of Chinese characters is conventionally categorized into the ``Seven Chinese Scripts'', developing continuously across successive dynasties. Originating in the Shang Dynasty, the Oracle Bone script represents the earliest mature writing system~\cite{OBI_Bench_2025, Divination_2008, Three_works_oracle_1997}. Carved on tortoise shells and animal bones, it features thin, angular lines, strong pictographic characteristics, and lacks standardized character sizes or spatial alignment.
During the Shang and Zhou Dynasties, the Bronze script emerged on ceremonial vessels, exhibiting thicker strokes that gradually evolved into more regularized and aesthetic structures by the late Western Zhou period~\cite{research_tool_bronze_2020, BIRD_2025, LadderMoE_2025}.
Following the unification of China, the Qin Dynasty standardized the Seal script, simplifying earlier variants into fixed structural patterns with pronounced curvilinear symmetry~\cite{chinese_seal_2026, Qin_seal_2024, small_seal_script_2023}.
The Han Dynasty then witnessed the ``Clerical Reformation'' through the Clerical script~\cite{Stroke_extraction_2022, calligraphy_image_2025, han_protrait_2024}. This script flattened characters and replaced curves with angular strokes, marking the transition of Chinese characters into symbolic forms and laying the foundation for modern typography.
Emerging in the late Han and Wei-Jin periods, the Regular script established strict square-shaped structures and standardized strokes, remaining the dominant formal script to this day~\cite{Stroke_chinese_2017, Evolution_fewshot_2022, TKH_MTH_2018, advantages_regular_2021}.
While these first five scripts successively served as formal writing systems, the Cursive and Running scripts developed primarily for informal and rapid writing~\cite{RS_GAN_2025}. Originating in the Han Dynasty and later evolving into unconstrained forms such as \textit{Kuangcao}, the Cursive script uses rapid, continuous strokes that often eliminate independent character boundaries~\cite{cursive_detection_2020, recognition_cursive_2020, regular_to_cursive_2018, unconstrained_cursive_1995, huai_su_cursive_1998, aesthetic_cursive_2023}. In contrast, the Running script functions as a fluid yet legible intermediate style.
From a computer vision perspective, this historical progression introduces substantial visual distribution shifts, as texts evolve from pictographic drawings to structured geometric symbols and further into highly abstract continuous forms~\cite{EVOBC_2024, Graph_evolution_2025, Evolution_fewshot_2022}. These shifts create significant challenges for text localization, cross-period feature mapping, and reading order prediction, often causing modern VLMs to hallucinate by overfitting to familiar contemporary glyph patterns rather than accurately interpreting ancient structures~\cite{Real_World_Doc_2026, MMTIT_2026, SENTINEL_2025}.

\subsection{VLLMs in Modern Text Perception}

With the rapid advancement of Large Language Models (LLMs)~\cite{GPT3_2020, GPT_4o_2023, GPT_5_2025}, VLLMs have achieved strong alignment between visual and textual representations through cross-modal integration, marking a major step toward general-purpose AI systems~\cite{Qwen2_VL_2024, Qwen_VL_2023, Qwen2_5_VL_2025, LLaVA_v1_2023, LLaVA_v1_5_2024, LLaVA_NeXT_2024, InstructBLIP_2023, Uni_DPO_2025, OpenAI_GPT4V_2023, MiniGPT_4_2024, hou2026uni}. Building on this progress, VLLMs have recently driven a paradigm shift in text-rich visual understanding, moving from cascaded OCR pipelines to end-to-end multimodal perception~\cite{PaddleOCR_VL_2025, HunyuanOCR_2025}. By combining high-resolution visual encoders with powerful language models, VLLMs directly align visual features with semantic spaces without requiring intermediate text-line cropping.
Pre-trained on massive image-text corpora, state-of-the-art models (\myeg GPT-4o~\cite{GPT_4o_2023}, Qwen-VL~\cite{Qwen3_VL_2025, Qwen2_5_VL_2025, Qwen3_5_2026}) demonstrate strong zero-shot performance on modern benchmarks such as OCRBench~\cite{OCRBench_v2_2024} and OmniDocBench~\cite{Omnidocbench_2025}. They effectively handle text localization, complex layout parsing, and cross-modal reasoning in real-world scenarios. However, this performance largely depends on the unified morphology, clear whitespace delimiters, and regular geometric structures of modern writing systems. Their visual encoders are primarily optimized for standardized contemporary typography (\myeg printed serif and sans-serif fonts) dominant in pre-training corpora. Consequently, when facing the extreme historical distribution shifts, unconstrained layouts, and morphological diversity described above, the perceptual limitations of VLLMs across historical scripts remain largely unexplored~\cite{TongGu_VL_2025, OracleAgent_2025, V_oracle_2025, WenyanGPT_2025}.

\input{tables/benchmark_comparison}

\subsection{Evaluation for Ancient Script Perception}

Existing research on ancient script perception suffers from two major limitations. First, most existing benchmarks focus only on specific historical periods or isolated script categories, resulting in a highly fragmented evaluation landscape. For example, recent works such as HWOBC~\cite{HWOBC_2020}, HUST-OBC~\citep{HUST_OBC_2024}, OBIMD~\citep{OBIMD_2024}, and OBI-Bench~\cite{OBI_Bench_2025} are dedicated exclusively to the multi-modal understanding of Oracle Bone Inscriptions. In contrast, document benchmarks such as M5HisDoc~\cite{M5HisDoc_2023} and HisDoc1B~\citep{HisDoc1B_2025} mainly target paragraph-level OCR in mature ancient Chinese books. Second, some recent studies, such as HWOBC~\cite{HWOBC_2020}, HUST-OBC~\cite{HUST_OBC_2024}, and GEVO-Bench~\cite{GEVO_Bench_2026}, only provide character-level recognition tasks for ancient scripts, which severely limits the evaluation of holistic document-level capabilities, including full-image recognition, structural parsing, contextual analysis, and semantic understanding.
While these datasets are valuable within their respective domains, they fail to provide a continuous evolutionary perspective spanning the entire development trajectory of Chinese scripts. Concurrently, as VLLMs emerge as universal perception engines, it becomes increasingly important to evaluate their robustness against the systematic visual distribution shifts accumulated across thousands of years of script evolution~\cite{OBI_Bench_2025, MCS_Bench_2025}. However, the current fragmentation of datasets makes such a holistic evaluation fundamentally unattainable.
To bridge this gap, \textbf{Chronicles-OCR} provides a unified benchmark specifically designed to establish performance baselines and quantify the perceptual bottlenecks of VLLMs across the complete evolutionary trajectory of Chinese scripts. By introducing stage-adaptive evaluation tasks, ranging from cross-period character spotting and fine-grained visual referring recognition to complex ancient text parsing and script classification, we enable the first holistic assessment of cross-temporal visual distribution shifts, thereby advancing the development of Digital Humanities.

%% file: tables/benchmark_comparison.tex
\begin{table}[th]
    \centering
    \captionsetup{font={small}}
    \caption{
        \textbf{Comparison of Ancient Chinese Script Benchmarks.}
        Chronicles-OCR provides the first unified benchmark covering the full evolutionary trajectory of the Seven Chinese Scripts with comprehensive task diversity.
    }
    \vspace{-1.0em}
    {
        \renewcommand{\arraystretch}{0.92}
        \setlength{\tabcolsep}{4pt}
        \resizebox{\columnwidth}{!}{
            \begin{tabular}{l c cccccccc ccc}
                \toprule
                \multirow{2}{*}[-2.8mm]
                {\textbf{Benchmark}}                          &
                \multirow{2}{*}[-2.8mm]
                {\textbf{\makecell{Release \nextline Date}}}  &
                \multicolumn{7}{c}{\textbf{Script Types}}     &
                \multicolumn{4}{c}{\textbf{Task Types}}
                \\
                \cmidrule(lr){3-9}
                \cmidrule(lr){10-13}
                                                              &
                                                              &
                \makecell{Oracle \nextline Bone} 
                                                              &
                \makecell{Bronze \nextline Script} 
                                                              &
                \makecell{Seal \nextline Script} 
                                                              &
                \makecell{Clerical \nextline Script} 
                                                              &
                \makecell{Regular \nextline Script} 
                                                              &
                \makecell{Cursive \nextline Script} 
                                                              &
                \makecell{Running \nextline Script} 
                                                              &
                \makecell{Character \nextline Spotting} 
                                                              &
                \makecell{Character \nextline Recognition} 
                                                              &
                \makecell{Text \nextline Parsing} 
                                                              &
                \makecell{Script \nextline Classification} 
                \\
                \midrule
                TKH \& MTH~\cite{TKH_MTH_2018}                & 2018.05 &        &        &        &        & \cmark &        &        & \cmark & \cmark & \cmark &        \\
                HWOBC~\cite{HWOBC_2020}                       & 2020.11 & \cmark &        &        &        &        &        &        & \cmark &        &        &        \\
                M5HisDoc~\cite{M5HisDoc_2023}                 & 2023.11 &        &        &        & \cmark & \cmark & \cmark & \cmark & \cmark & \cmark & \cmark &        \\
                HUST-OBC~\cite{HUST_OBC_2024}                 & 2024.01 & \cmark &        &        &        &        &        &        & \cmark &        &        &        \\
                EVOBC~\cite{EVOBC_2024}                       & 2024.01 & \cmark & \cmark & \cmark & \cmark &        &        &        & \cmark &        &        &        \\
                OBI Component 20~\cite{OBI_Component_20_2024} & 2024.06 & \cmark &        &        &        &        &        &        & \cmark &        &        &        \\
                OBIMD~\cite{OBIMD_2024}                       & 2024.07 & \cmark &        &        &        &        &        &        & \cmark & \cmark & \cmark &        \\
                OBI-Bench~\cite{OBI_Bench_2025}               & 2024.12 & \cmark &        &        &        &        &        &        & \cmark & \cmark &        &        \\
                HisDoc1B~\cite{HisDoc1B_2025}                 & 2025.01 &        &        &        & \cmark & \cmark & \cmark & \cmark & \cmark & \cmark & \cmark &        \\
                RMOBS~\cite{OracleFusion_2025}                & 2025.06 & \cmark &        &        &        &        &        &        & \cmark &        &        &        \\
                PictOBI-20k~\cite{Pictobi_2026}               & 2025.09 & \cmark &        &        &        &        &        &        & \cmark &        &        &        \\
                GEVO-Bench~\cite{GEVO_Bench_2026}             & 2026.04 & \cmark & \cmark & \cmark & \cmark & \cmark &        &        & \cmark &        &        & \cmark \\
                \midrule
                \rowcolor{table_ours}
                \textbf{Chronicles-OCR}                       & 2026.05 & \cmark & \cmark & \cmark & \cmark & \cmark & \cmark & \cmark & \cmark & \cmark & \cmark & \cmark \\
                \bottomrule
            \end{tabular}
        }
    }
    \vspace{-10pt}
    \label{tab:ancient_benchmark_comparison}
\end{table}

%% file: section/3_benchmark.tex
\section{Chronicles-OCR Benchmark}
\label{sec:benchmark}

\begin{figure*}[t]
    \centering
    \includegraphics[width=\linewidth]{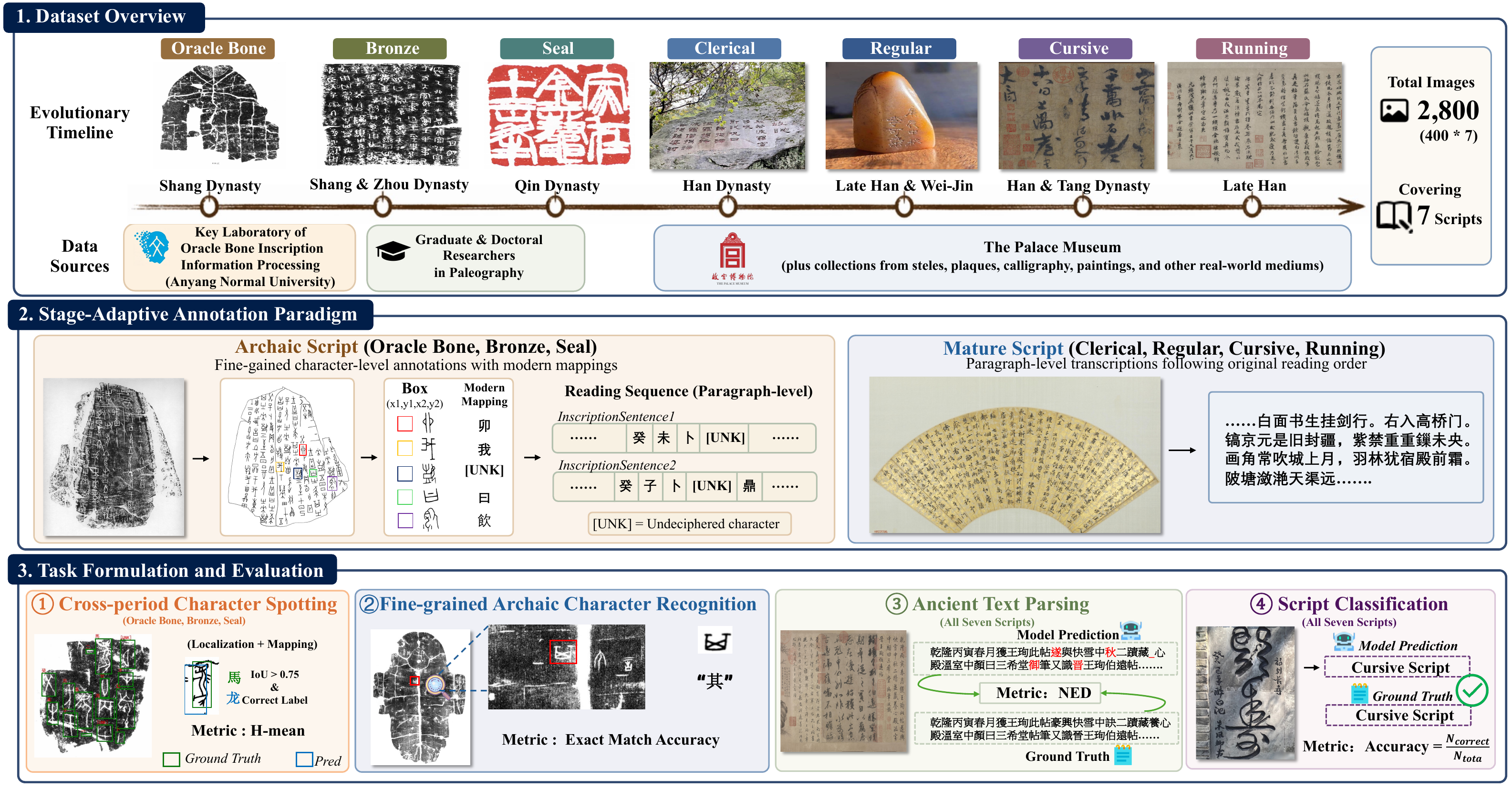}
    \vspace{-1.8em}
    \captionsetup{font={small}}
    \caption{
        \textbf{Overview of the Chronicles-OCR Benchmark.}
        The benchmark integrates three core components: (1) \textbf{Data Curation and Image Sourcing} across the Seven Chinese Scripts' evolutionary timeline; (2) \textbf{Stage-Adaptive Annotation Paradigm}, applying character-level grounding for archaic scripts and sequence-level transcriptions for mature ones; and (3) \textbf{Task Formulation}, establishing four differentiated evaluation tasks to quantify the perceptual bottleneck of VLLMs.
    }
    \label{fig:chronocr}
\end{figure*}

In this section, we detail the construction of the Chronicles-OCR benchmark. As illustrated in~\cref{fig:chronocr}, Chronicles-OCR is organized into three core components. We first introduce the expert-driven image sourcing and data curation process in~\cref{subsec: data_curation}.
This is followed by a detailed exposition of the \textit{Stage-Adaptive Annotation Paradigm} in~\cref{subsec: data_annotation}, which is specifically tailored to address the morphological variations of scripts across different historical eras.
Finally, we formulate four rigorous evaluation tasks and corresponding metrics to precisely quantify the perceptual bottlenecks of current VLLMs in~\cref{subsec: evaluation_metrics}. As summarized in~\cref{tab:ancient_benchmark_comparison}, compared to existing datasets, Chronicles-OCR stands as the first unified benchmark to encompass the full evolutionary trajectory of the Seven Chinese Scripts while offering unprecedented task diversity.

\subsection{Image Sourcing and Data Curation}
\label{subsec: data_curation}

Securing highly reliable historical image data is paramount, as it forms the foundational basis of the Chronicles-OCR benchmark.
To ensure historical authenticity and structural diversity, our raw image sourcing process was conducted in close collaboration with domain experts and institutional partners.
Specifically, the Oracle Bone image data was provided by the Key Laboratory of Oracle Bone Inscription Information Processing (Anyang Normal University). The Bronze and Seal script images were systematically curated by doctoral and graduate researchers specializing in paleography. Furthermore, the construction of the Clerical, Regular, Running, and Cursive script datasets involved a collaborative effort with the Palace Museum. While a substantial portion of the core image data was sourced directly from the museum's archives, the dataset was further enriched by digitizing samples from diverse real-world physical media, including historical plaques, stone steles, calligraphic scrolls, and ancient paintings.

\subsection{Stage-Adaptive Annotation Paradigm}
\label{subsec: data_annotation}

To guarantee the highest standard of ground-truth quality, the entire annotation process was rigorously conducted and cross-verified by a multi-tier team of domain experts, meticulously matched to specific historical scripts. Specifically, the annotations for Oracle Bone inscriptions were executed and reviewed by researchers from the Key Laboratory of Oracle Bone Inscription Information Processing at Anyang Normal University. For the Bronze and Seal scripts, the annotation tasks were undertaken by specialized Master's and Ph.D. scholars with profound expertise in paleography and ancient Chinese philology. Furthermore, the mature scripts, comprising the Clerical, Regular, Running, and Cursive styles, were curated and verified by experts from the Palace Museum. This rigorous, expert-driven pipeline ensures the exceptional fidelity in script classification, bounding box localization, and character transcription. Crucially, built upon this authoritative foundation, we designed a \textit{Stage-Adaptive Annotation Paradigm} tailored to the distinct evolutionary characteristics of these scripts.

For archaic scripts (Oracle Bone, Bronze, Seal), the texts differ drastically from modern Chinese not only in spatial layout but also in fundamental morphological structures. Furthermore, due to their extreme historical antiquity, the physical media (\myeg tortoise shells, weathered bronzes) introduce severe background noise and degradation. Relying solely on sequence-level transcription for these stages is fundamentally insufficient, as it conflates spatial layout confusion with character-level decipherment failure. To rigorously decouple these challenges and ensure a precise evaluation of VLLMs, we exclusively provide fine-grained, character-level annotations for these three archaic scripts.
Specifically, the annotations consist of single-character bounding boxes to localize unconstrained symbols, alongside modern character mappings to bridge the profound semantic gap.
For ancient characters that remain undeciphered by modern paleography, we uniformly annotate them with a special \texttt{[UNK]} token. Finally, we provide paragraph-level annotations organized strictly according to the original reading sequences.

Conversely, for mature pre-modern scripts (Clerical, Regular, Running, Cursive), we adopt line- and paragraph-level transcriptions. The rationale is that scripts in these mature stages typically appear in continuous paragraph formats, possess high inter-character discriminability, and share fundamental topological structures with modern Chinese characters. Consequently, forcing character-level bounding boxes becomes redundant and sometimes counterintuitive (especially for continuous Cursive strokes). For these stages, the evaluation focus naturally shifts to sequence-level continuous recognition.

\subsection{Task Formulation and Evaluation Metrics}
\label{subsec: evaluation_metrics}

Our task formulation is deeply rooted in the vision of empowering Digital Humanities with VLLMs. Based on the curated dataset and stage-adaptive annotations, we design four evaluation tasks that not only establish rigorous perceptual baselines but also closely mirror the practical scenarios in paleographic research and historical archiving:

\begin{itemize}[
        label=\raisebox{0.5ex}{\tiny$\bullet$},
        leftmargin=1em,
        itemsep=2pt, 
        parsep=0pt, 
        topsep=0pt, 
        partopsep=0pt 
    ]
    \item \textbf{Cross-period Character Spotting:} As illustrated in the first block of the bottom panel in~\cref{fig:chronocr}, this end-to-end task is designed to assist paleographic researchers with automated pre-annotation, assessing the model's dual capacity for visual grounding and morphological decipherment on unconstrained artifacts (e.g., raw bone rubbings). Evaluated exclusively on the Oracle Bone, Bronze, and Seal scripts, the VLLM is required to simultaneously output the bounding box coordinates and the corresponding modern Chinese character mapping for all archaic symbols present in an image. A prediction is deemed a True Positive (TP) if its Intersection over Union (IoU) with the ground truth reaches a strict threshold of $0.75$ and it exactly matches the mapped modern character. Undeciphered symbols annotated as \texttt{[UNK]} are strictly \textit{excluded} from the evaluation pool. Performance is evaluated using H-mean, the standard text spotting metric.

    \item \textbf{Fine-grained Archaic Character Recognition:}  As depicted in the second block of the bottom panel in~\cref{fig:chronocr}, when experts seek AI assistance for an ambiguous symbol on a rubbing, they intuitively point to the visual region rather than inputting numerical coordinates. To reflect this interactive expert-in-the-loop scenario and explicitly isolate character-level decipherment capabilities from spatial grounding, we introduce a \textit{visual referring} mechanism for archaic scripts. The target symbol is highlighted with a distinct colored bounding box directly on the input image. The model is then prompted (\myeg ``\textit{Recognize the archaic character highlighted by the red box in the image}'') to generate the corresponding modern character. This allows us to rigorously quantify pure morphological mapping accuracy. Symbols annotated as \texttt{[UNK]} are never sampled for this task. The performance is measured by the Exact Match Accuracy of the recognized characters.

    \item \textbf{Ancient Text Parsing:} As shown in the third block of the bottom panel in~\cref{fig:chronocr}, this task is oriented toward general digital transcription scenarios, assessing whether the model can comprehend historical spatial layouts (e.g., right-to-left, top-to-bottom columns) and correctly transcribe the text strictly along the original reading sequence. Evaluated across all seven scripts, we utilize the paragraph-level Normalized Edit Distance (NED) score, formulated as:

          \vspace{-1.0em}
          \begin{equation}
              \text{NED} = 1 - \frac{D(s_{\text{pred}}, s_{\text{gt}})}{\max(|s_{\text{pred}}|, |s_{\text{gt}}|)},
              \label{eq: NED_score}
          \end{equation}
          \par\nobreak\vspace{-0.5em}

          where $D(\cdot, \cdot)$ represents the Levenshtein edit distance~\cite{Levenshtein_1966}, and $s_{\text{pred}}, s_{\text{gt}}$ denote the predicted and ground-truth sequences, respectively. Similar to the spotting task, all \texttt{[UNK]} tokens are filtered out from both sequences before calculating the edit distance. This metric strictly penalizes any sequence-ordering mismatches, thereby accurately reflecting the model's layout comprehension without being conflated with decipherment issues.

    \item \textbf{Script Classification:} As presented in the fourth block of the bottom panel in~\cref{fig:chronocr}, this novel task is oriented toward automated document cataloging and archival sorting, probing the VLLM's macro-level understanding of morphological evolution. Given an image, the model must classify it into one of the ``Seven Chinese Scripts''. Spanning the entire dataset, this standard classification task is evaluated using overall Accuracy (Acc), defined as $\text{Acc} = \frac{N_{correct}}{N_{total}}$, where $N_{correct}$ is the number of accurately classified images.
\end{itemize}

%% file: section/4_experiments.tex
\section{Experiments}
\label{sec:experiments}

\subsection{Experimental Setup}
To comprehensively assess the perceptual boundaries of current visual-language models under historical distribution shifts, we assemble an extensive evaluation suite comprising a wide spectrum of state-of-the-art VLLMs. This suite covers both industry-leading proprietary models evaluated via official API endpoints and powerful open-source foundation models with publicly available weights, ensuring a rigorous and comprehensive comparison. As detailed in~\cref{subsec: evaluation_metrics}, all models are subjected to our unified benchmark protocol across four core tasks. Specifically, the Cross-period Character Spotting task evaluates fine-grained localization and morphological decipherment on Archaic Scripts, quantified by the standard text spotting H-mean score. The Fine-grained Archaic Character Recognition task leverages interactive visual referring to assess pure character-level mapping accuracy, evaluated via Exact Match. The Ancient Text Parsing task assesses the ability to comprehend and organize reading sequences, measured by the Normalized Edit Distance (NED). Finally, the Script Classification task tests macro-level feature perception, evaluated via standard classification Accuracy.

\input{tables/archaic_results}

\input{tables/mature_results}

\subsection{Main Results and Analysis}

\cref{tab:archaic_results,tab:mature_results} present the quantitative results across all evaluation tracks for Archaic and Mature scripts, respectively.

\input{figure_code/spotting_vis}

\mypara{Performance on Cross-period Character Spotting.}
The results on this task expose the most critical vulnerability of current VLLMs: a severe, twofold bottleneck in fine-grained grounding and morphological decipherment. For archaic unstandardized texts (Oracle Bone, Bronze, Seal), the vast majority of models almost entirely fail at this end-to-end task. Leading commercial models such as GPT-5~\cite{GPT_5_2025} and Gemini 2.5 Pro~\cite{Gemini_2_5_2025} register Spotting H-mean scores near zero. While the Seed2.0 Pro~\cite{Seed_2_0_2026} model demonstrates a relative advantage (achieving a Spotting H-mean of 21.1), the absolute performance remains exceptionally low. As qualitatively visualized in~\cref{fig:spotting_vis}, this catastrophic degradation is compounded by two distinct failures. First, in terms of spatial localization accuracy, VLLMs fundamentally lack the robust grounding mechanisms needed to isolate unconstrained, highly variable symbols embedded in noisy physical media (\myeg weathered bronzes and cracked tortoise shells). They rely heavily on modern layout priors, which completely break down on ancient artifacts, resulting in severe \textit{missed detections} and \textit{hallucinations} over background noise. Second, in pure character recognition, even if an archaic symbol is successfully localized, models still face a profound barrier to decipherment. The severe morphological deviation from early pictographic glyphs to modern abstract strokes creates a massive semantic gap, leading to frequent \textit{recognition errors}. Without explicit paleographic alignment, VLLMs fail to map these ancient historical morphologies to their modern counterparts, ultimately driving the End-to-End Spotting H-mean toward zero. Consequently, even the most capable leading models remain far from meeting the practical expectations of providing minimally viable automated pre-annotations to assist paleographic researchers in the Digital Humanities.

\input{figure_code/fine_grain_vis}

\mypara{Performance on Fine-grained Archaic Character Recognition.}
Evaluated using Exact Match via the interactive visual referring mechanism, this task explicitly isolates morphological decipherment from spatial grounding. As shown in~\cref{tab:archaic_results}, relieving models of the localization burden yields a relative performance uplift. For instance, Kimi K2.5 improves from a 6.6 Spotting score to 31.9 in Fine-grained Recognition. However, the absolute accuracy remains strikingly low, peaking at only 31.9\% on average and plummeting to 14.0\% (Gemini 3.1 Pro) on the earliest Oracle Bone Script. As qualitatively illustrated in~\cref{fig:fine_grain_vis}, even when explicitly prompted with bounding boxes highlighting the exact ancient symbol, models consistently fail to map these pictographic glyphs to their modern counterparts. This confirms our second hypothesis: beyond spatial layout confusion, there exists a massive, independent semantic gap. Current VLLMs fundamentally lack the specialized paleographic representations required to decode unstandardized historical morphologies, leading to persistent recognition errors.

\input{figure_code/parsing_vis}

\mypara{Performance on Ancient Text Parsing.}
Evaluated via NED, this task reveals a clear and significant performance gap between mature and archaic scripts. Models exhibit relatively stronger parsing capabilities at mature stages. For example, Kimi K2.5~\cite{Kimi_K2_5_2026} achieves a NED score of 0.78 on Regular Script, which closely resembles modern printed text. While this score indicates partial comprehension, it still falls short of the near-perfect parsing typical in modern document OCR. As qualitatively visualized in~\cref{fig:parsing_vis}, even on mature scripts with standard reading conventions, models frequently produce severe transcription errors and hallucinations when decoding complex strokes. More alarmingly, its performance drops substantially to 0.19 on Bronze Script and plummets to merely 0.06 on Oracle Bone Script. This massive degradation stems from two primary factors. First, the severe morphological deviation of archaic characters from modern Chinese makes the individual symbols inherently difficult for VLLMs to comprehend. Second, there is a fundamental difference in layout structures. Mature scripts (Clerical, Regular, Running, Cursive) generally adhere to standardized reading conventions (\myeg continuous vertical columns). Conversely, archaic inscriptions often feature highly unconstrained, non-linear reading sequences scattered randomly across physical artifacts. These results indicate that current VLLMs rely heavily on both the familiar character shapes and the rigid layout priors of modern documents, struggling to logically organize text when these fundamental rules are broken.

\input{figure_code/class_vis}

\mypara{Performance on Script Classification.}
As visualized in~\cref{fig:class_vis}, an intriguing contradiction emerges when analyzing the Script Classification results across the historical timeline. Surprisingly, models achieve exceptionally high classification accuracy on Archaic Scripts (\myeg Seed2.0 Pro~\cite{Seed_2_0_2026} at 95.5\%, and Kimi K2.5~\cite{Kimi_K2_5_2026} at 95.2\%). However, this macro-level success sharply contrasts with their performance on Mature Scripts, where classification accuracy drops significantly (\myeg both Seed2.0 Pro and Kimi K2.5 fall to 78.1\%). This inversion highlights a fundamental decoupling between stylistic recognition and fine-grained perception. Just as a human might visually identify a text as Arabic without comprehending its meaning, VLLMs demonstrate a strong capacity to recognize the global morphological style of ancient scripts or the distinct contextual textures of their physical mediums (\myeg tortoise shells or bronze artifacts). They can confidently categorize archaic scripts based on these macro-level visual priors. Conversely, mature scripts (Clerical, Regular, Running, Cursive) are predominantly written on identical physical mediums (ink on paper) and share a unified topological framework. Distinguishing among them requires perceiving subtle stroke dynamics, such as the degree of cursiveness or specific brush connections. Since VLLMs lack micro-level character perception, as evidenced by their failure in the Spotting task, they struggle to capture these delicate stroke variations, leading to severe confusion among mature script categories.

\subsection{Insights and Implications}

\mypara{The Dependence of Reasoning on Foundational Perception.}
As reflected by the overall trends in~\cref{tab:archaic_results,tab:mature_results}, scaling up model parameters consistently yields better overall performance. However, an unexpected phenomenon emerges when analyzing reasoning-enhanced (``think'') model variants: enabling explicit reasoning generally leads to performance degradation across both mature and archaic scripts. Our qualitative observations suggest that the generated reasoning processes are often redundant, irrelevant, or erroneous, introducing additional hallucinations instead of correcting perception mistakes. This indicates that current reasoning mechanisms in VLLMs remain highly dependent on reliable visual perception. When the perceptual foundation is unstable, explicit reasoning may amplify uncertainty and turn tentative recognition errors into highly confident but incorrect predictions.

\mypara{The Gap in Fine-grained Feature Mining.}
The widespread failure of VLLMs in mature script classification offers a subtle but critical insight. It suggests that current visual-language perception of text may still be confined to macroscopic shape recognition, with a long way to go for perceiving deeper, fine-grained characteristics. Distinguishing mature script categories necessitates the perception of delicate stroke connections and specific stylistic dynamics, which current models struggle to capture accurately. This highlights a significant avenue for future research: shifting from rough structural perception toward mining fine-grained, stroke-level features.

\mypara{Bridging AI and Digital Humanities.}
The severe performance degradation on archaic scripts confirms that VLLMs still face a long and challenging journey in the realm of Digital Humanities. Currently, as reflected by their substantial failures in the Cross-period Character Spotting and Fine-grained Archaic Character Recognition tasks, these models remain far from the initial expectation of serving as reliable pre-annotation or interactive assistance tools for paleographers. Looking further ahead, the ultimate vision for AI in this domain is to assist experts in deciphering currently undeciphered ancient characters. By exposing these critical vulnerabilities, Chronicles-OCR highlights the immense cultural value of historical text perception, encouraging the community to build tools that can genuinely preserve and decode human historical heritage.

%% file: tables/archaic_results.tex
\begin{table*}[ht]
    \centering
    \captionsetup{font={small}}
    \caption{
        \textbf{Evaluation Results on Archaic Scripts (Oracle Bone, Bronze, Seal).}
        Performance across four core tasks: Character Spotting (Spot.), Fine-grained Recognition (Fine.), Ancient Text Parsing (Pars.), and Script Classification (Class.). \textbf{Bold} indicates the best performance while \ul{underlined} results denote the second-best performance.
    }
    \vspace{-0.8em}
    {
        \resizebox{\textwidth}{!}{
            \begin{tabular}{l c | cccc | cccc | cccc | cccc}
                \toprule
                \multirow{2}{*}[-1.0mm]
                {\textbf{Model}}                                         &
                \multirow{2}{*}[-1.0mm]
                {\textbf{Think}}                                         &
                \multicolumn{4}{c|}
                {\textbf{Average}}                                       &
                \multicolumn{4}{c|}
                {\textbf{Oracle Bone Script}}                            &
                \multicolumn{4}{c|}
                {\textbf{Bronze Script}}                                 &
                \multicolumn{4}{c}
                {\textbf{Seal Script}}                                                                                                                                                                                                                                                                                                                                \\
                \cmidrule(lr){3-6}
                \cmidrule(lr){7-10}
                \cmidrule(lr){11-14}
                \cmidrule(lr){15-18}
                                                                         &        & \textbf{Spot.} & \textbf{Fine.} & \textbf{Pars.} & \textbf{Class.} & \textbf{Spot.} & \textbf{Fine.} & \textbf{Pars.} & \textbf{Class.} & \textbf{Spot.} & \textbf{Fine.} & \textbf{Pars.} & \textbf{Class.} & \textbf{Spot.} & \textbf{Fine.} & \textbf{Pars.} & \textbf{Class.} \\
                \midrule
                \multicolumn{18}{c}{\emph{\quad \textbf{Open-Source Models}}}                                                                                                                                                                                                                                                                                         \\
                \logo{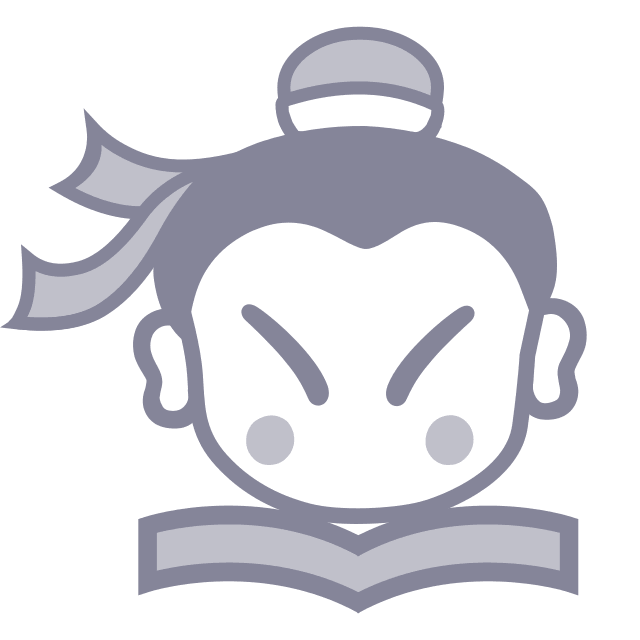}InternVL3.5-8B~\cite{InternVL3_5_2025}      &        & 0.1            & 5.9            & 0.07           & 56.6            & 0.0            & 1.0            & 0.01           & 85.8            & 0.0            & 2.2            & 0.03           & 7.0             & 0.2            & 14.5           & 0.17           & 77.0            \\
                \logo{intern}InternVL3.5-A28B~\cite{InternVL3_5_2025}    &        & 0.5            & 15.7           & 0.13           & 79.0            & 0.0            & 2.5            & 0.02           & 96.3            & 0.4            & 7.8            & 0.08           & 79.2            & 1.0            & 36.8           & 0.29           & 61.5            \\
                \logo{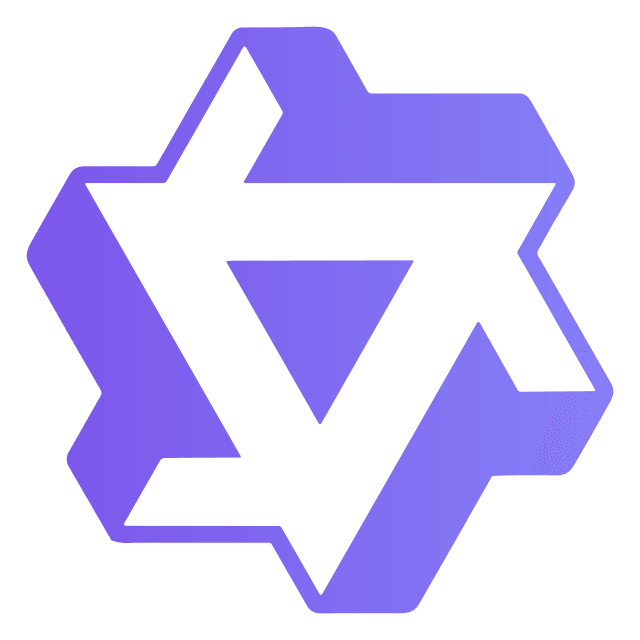}Qwen2.5-VL-7B~\cite{Qwen2_5_VL_2025}          &        & 0.0            & 7.4            & 0.07           & 71.8            & 0.0            & 4.0            & 0.03           & 93.8            & 0.0            & 4.5            & 0.04           & 22.5            & 0.0            & 13.8           & 0.14           & \ul{99.2}       \\
                \logo{qwen}Qwen2.5-VL-72B~\cite{Qwen2_5_VL_2025}         &        & 0.0            & 0.0            & 0.07           & 74.2            & 0.0            & 0.0            & 0.01           & 98.0            & 0.0            & 0.0            & 0.04           & 26.0            & 0.0            & 0.0            & 0.16           & 98.5            \\
                \logo{qwen}Qwen3-VL-2B~\cite{Qwen3_VL_2025}              &        & 2.1            & 10.7           & 0.12           & 73.0            & 0.0            & 1.4            & 0.00           & 96.6            & 0.8            & 6.8            & 0.06           & 36.5            & 5.7            & 24.0           & 0.31           & 85.8            \\
                \logo{qwen}Qwen3-VL-8B~\cite{Qwen3_VL_2025}              &        & 3.4            & 17.3           & 0.18           & 73.7            & 0.2            & 3.4            & 0.01           & 98.6            & 2.5            & 11.0           & 0.10           & 24.0            & 7.5            & 37.5           & 0.42           & 98.5            \\
                \logo{qwen}Qwen3-VL-8B~\cite{Qwen3_VL_2025}              & \cmark & 1.0            & 9.1            & 0.10           & 67.3            & 0.0            & 3.7            & 0.04           & 97.7            & 0.2            & 7.0            & 0.05           & 31.8            & 2.8            & 16.8           & 0.20           & 72.5            \\
                \logo{qwen}Qwen3-VL-A22B~\cite{Qwen3_VL_2025}            &        & 7.8            & 17.5           & 0.19           & 91.8            & 0.3            & 5.4            & 0.02           & 99.2            & 6.5            & 12.2           & 0.12           & 80.2            & 16.6           & 35.0           & 0.43           & 96.0            \\
                \logo{qwen}Qwen3-VL-A22B~\cite{Qwen3_VL_2025}            & \cmark & 2.1            & 13.6           & 0.17           & 87.3            & 0.1            & 4.2            & 0.04           & 98.0            & 0.9            & 10.2           & 0.11           & 66.8            & 5.3            & 26.2           & 0.38           & 97.2            \\
                \logo{qwen}Qwen3.5-A3B~\cite{Qwen3_5_2026}               &        & 5.6            & 16.2           & 0.20           & 76.5            & 0.2            & 5.1            & 0.03           & 99.7            & 5.3            & 11.5           & 0.12           & 30.0            & 11.2           & 32.0           & 0.45           & \textbf{99.8}   \\
                \logo{qwen}Qwen3.5-A17B~\cite{Qwen3_5_2026}              &        & 9.7            & 22.6           & 0.22           & 88.3            & 0.5            & 9.1            & 0.03           & 99.7            & 9.2            & 17.5           & 0.13           & 67.2            & 19.4           & 41.3           & 0.50           & 98.0            \\
                \logo{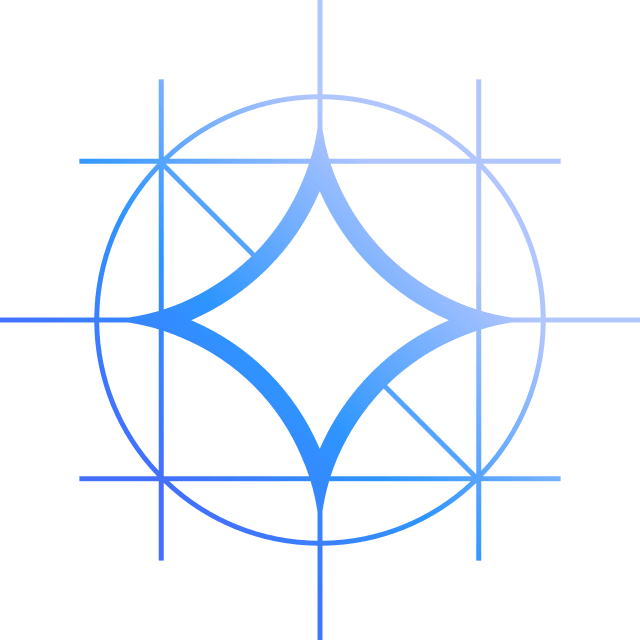}Gemma 4 31B it~\cite{Gemma_2024}             &        & 2.3            & 7.0            & 0.05           & 70.0            & 0.0            & 3.1            & 0.01           & 72.6            & 1.0            & 6.5            & 0.03           & 74.8            & 6.0            & 11.2           & 0.10           & 62.7            \\
                \logo{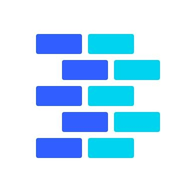}MiniCPM-V 4.5~\cite{MiniCPM_V_4_5_2025}    & \cmark & 0.0            & 5.7            & 0.03           & 65.2            & 0.0            & 2.5            & 0.01           & 95.2            & 0.0            & 5.5            & 0.03           & 18.0            & 0.1            & 9.0            & 0.04           & 82.5            \\
                \logo{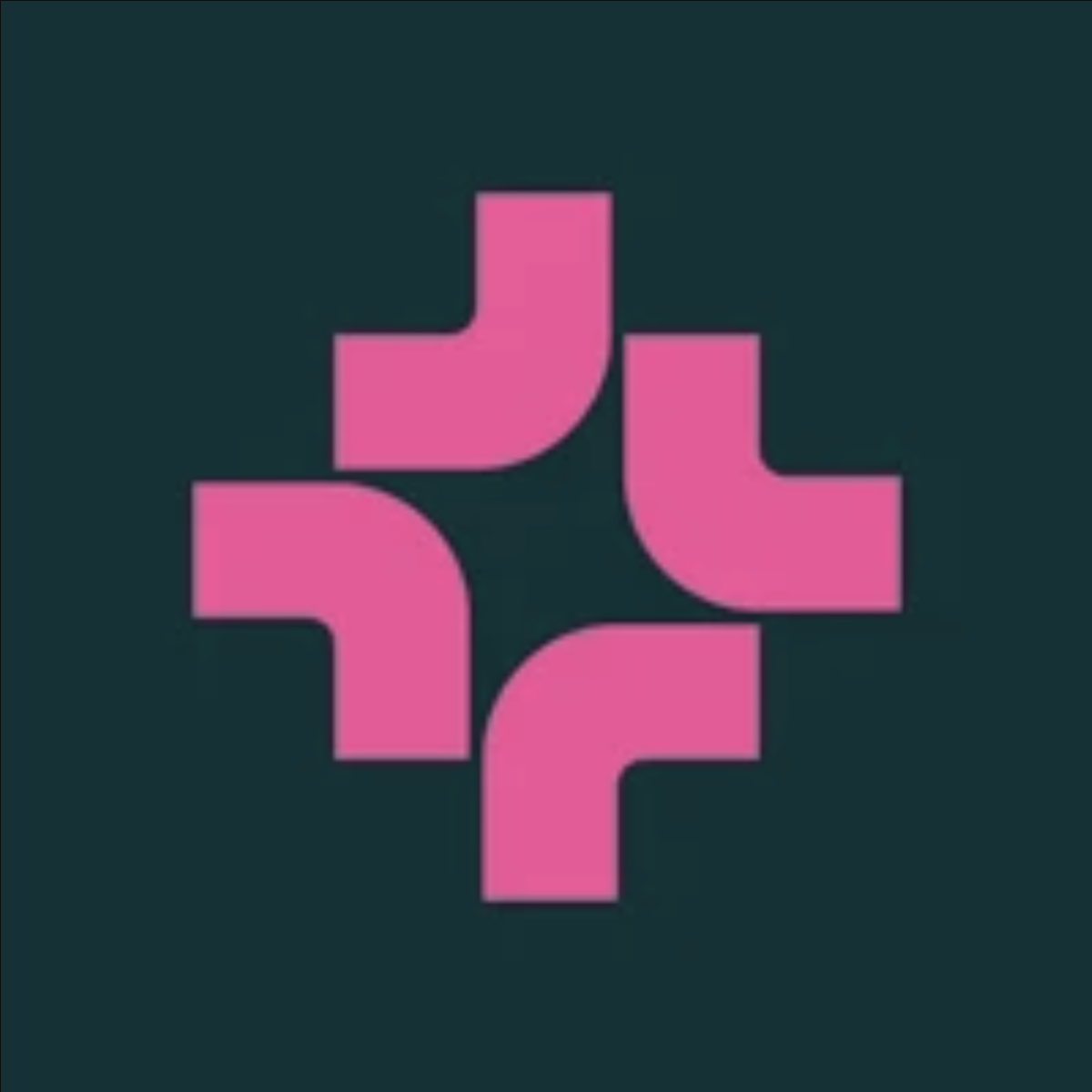}Molmo 7B-D 0924~\cite{Molmo_PixMo_2025}    &        & 0.0            & 0.1            & 0.00           & 20.4            & 0.0            & 0.0            & 0.01           & 40.8            & 0.0            & 0.2            & 0.00           & 0.0             & 0.0            & 0.0            & 0.00           & 20.5            \\
                \logo{AllenAI}Molmo 72B 0924~\cite{Molmo_PixMo_2025}     &        & 0.0            & 0.3            & 0.00           & 36.9            & 0.0            & 0.5            & 0.00           & 28.0            & 0.0            & 0.5            & 0.00           & 0.8             & 0.0            & 0.0            & 0.00           & 82.0            \\
                \logo{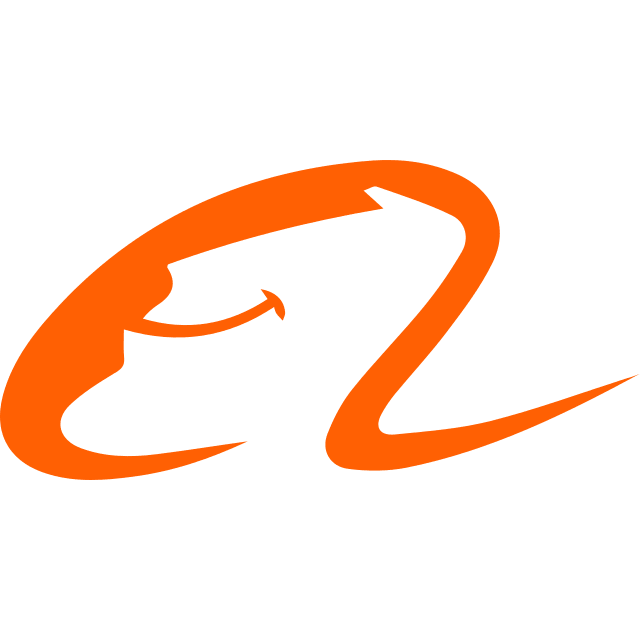}Ovis2.6-30B-A3B~\cite{Ovis2_5_2025}        & \cmark & 2.5            & 11.3           & 0.11           & 60.8            & 0.1            & 2.0            & 0.02           & 89.8            & 0.7            & 7.5            & 0.06           & 13.5            & 6.8            & 24.5           & 0.25           & 79.0            \\
                \logo{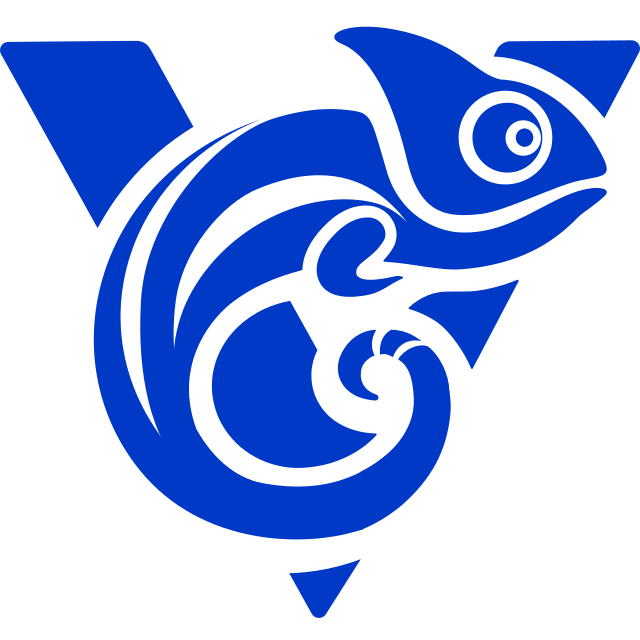}GLM-4.5V 108B~\cite{GLM_4_5V_2025}            & \cmark & 1.8            & 6.7            & 0.06           & 69.0            & 0.1            & 4.2            & 0.04           & \textbf{100}    & 2.0            & 6.5            & 0.05           & 15.5            & 3.3            & 9.2            & 0.10           & 91.5            \\
                \logo{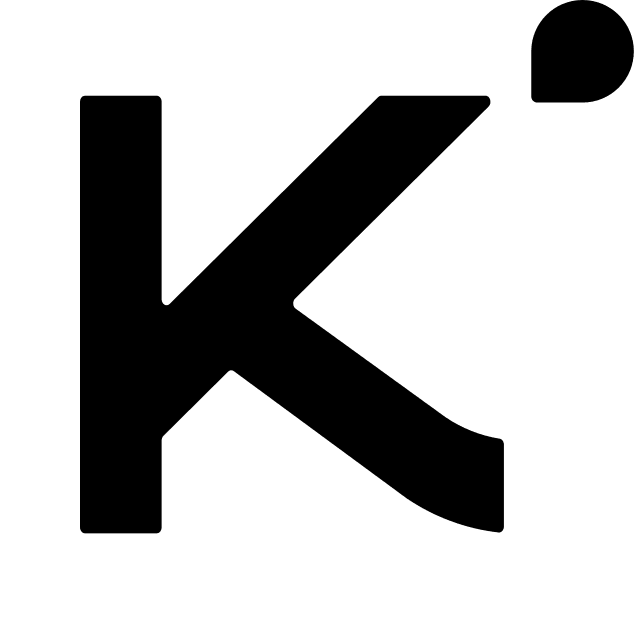}Kimi K2.5~\cite{Kimi_K2_5_2026}               &        & 6.6            & \textbf{31.9}  & \textbf{0.29}  & 95.2            & 0.1            & \ul{11.5}      & \ul{0.06}      & \textbf{100}    & 7.5            & 25.8           & 0.19           & \ul{90.0}       & 12.5           & \textbf{58.5}  & \textbf{0.60}  & 95.5            \\
                \logo{kimi}Kimi K2.5~\cite{Kimi_K2_5_2026}               & \cmark & 2.4            & 24.2           & \ul{0.28}      & 93.0            & 0.0            & 10.2           & \textbf{0.07}  & \ul{99.8}       & 1.2            & 17.5           & 0.20           & 85.8            & 6.0            & 44.8           & \ul{0.57}      & 93.5            \\
                \midrule
                \multicolumn{18}{c}{\emph{\quad \textbf{Proprietary Models}}}                                                                                                                                                                                                                                                                                         \\
                \logo{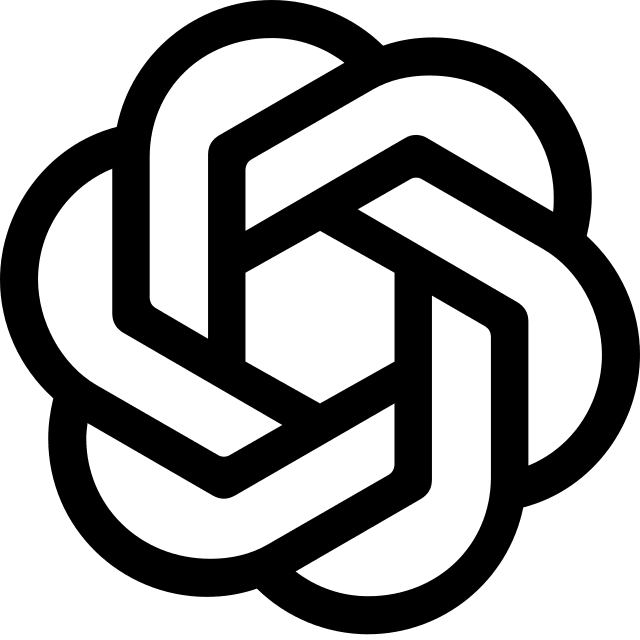}GPT-4o~\cite{GPT_4o_2023}                   &        & 0.1            & 2.0            & 0.03           & 77.4            & 0.0            & 0.5            & 0.01           & 96.5            & 0.0            & 1.0            & 0.02           & 46.8            & 0.3            & 4.5            & 0.06           & 89.0            \\
                \logo{openai}GPT-5~\cite{GPT_5_2025}                     &        & 0.5            & 4.2            & 0.06           & 85.4            & 0.0            & 4.0            & 0.01           & 98.2            & 0.0            & 4.0            & 0.04           & 60.5            & 1.6            & 4.5            & 0.12           & 97.5            \\
                \logo{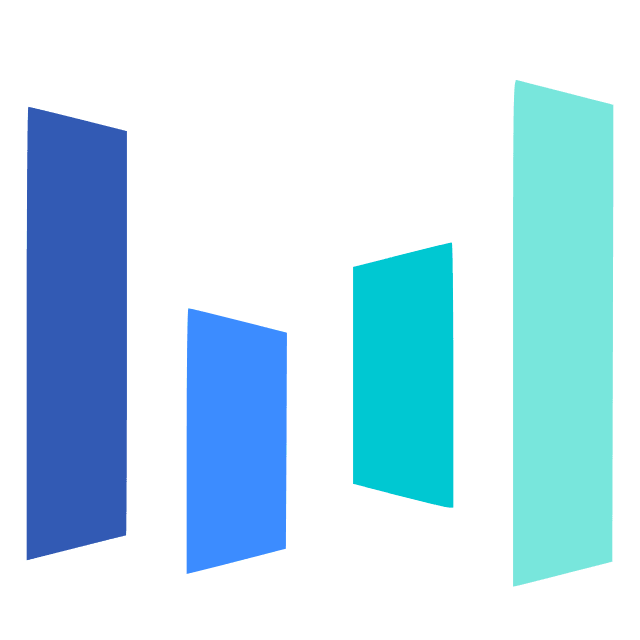}Seed1.8~\cite{Seed_1_8_2026}             &        & 11.9           & 23.3           & 0.21           & 93.0            & 0.4            & 9.2            & 0.04           & 99.5            & 9.4            & 15.8           & 0.17           & 80.5            & 26.7           & \ul{45.0}      & 0.42           & 99.0            \\
                \logo{bytedance}Seed1.8~\cite{Seed_1_8_2026}             & \cmark & 9.8            & 19.9           & 0.22           & \textbf{95.7}   & 0.4            & 8.8            & 0.05           & 99.5            & 5.8            & 14.8           & 0.18           & \ul{90.0}       & 23.3           & 36.2           & 0.43           & 97.5            \\
                \logo{bytedance}Seed2.0 Pro~\cite{Seed_2_0_2026}         &        & \textbf{21.1}  & \ul{27.7}      & 0.23           & 95.2            & \textbf{3.0}   & 11.0           & 0.04           & 99.5            & \textbf{19.9}  & \textbf{30.8}  & \ul{0.22}      & \textbf{92.2}   & \textbf{40.7}  & 41.5           & 0.43           & 93.8            \\
                \logo{bytedance}Seed2.0 Pro~\cite{Seed_2_0_2026}         & \cmark & \ul{19.8}      & 24.9           & 0.27           & \ul{95.5}       & \ul{2.4}       & 11.2           & 0.05           & \ul{99.8}       & \ul{17.8}      & \ul{26.0}      & \textbf{0.26}  & \textbf{92.2}   & \ul{39.1}      & 37.5           & 0.49           & 94.5            \\
                \logo{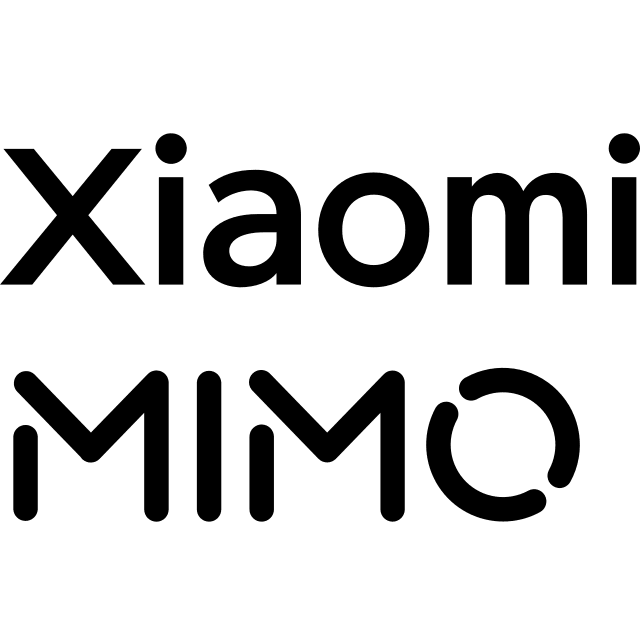}MiMo-V2-Omni~\cite{Mimo_V2_Omni_2026}   & \cmark & 0.6            & 8.1            & 0.09           & 83.7            & 0.0            & 6.5            & 0.05           & 99.5            & 0.2            & 8.0            & 0.07           & 58.5            & 1.5            & 9.8            & 0.15           & 93.0            \\
                \logo{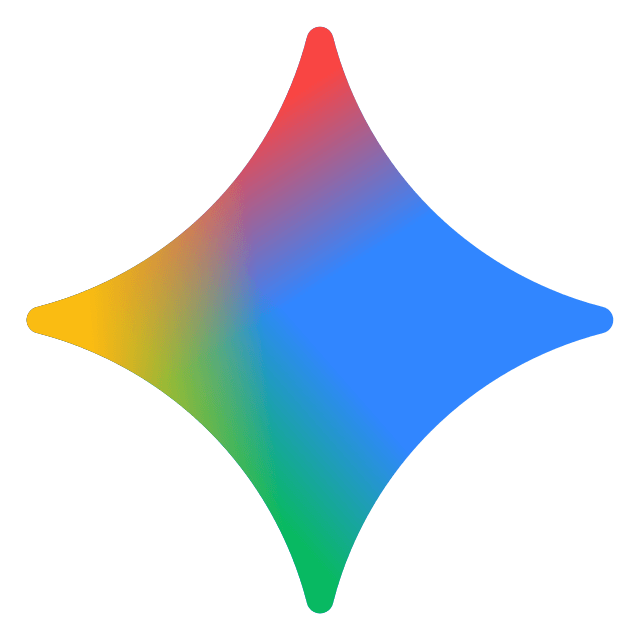}Gemini 2.5 Pro~\cite{Gemini_2_5_2025}       & \cmark & 1.0            & 7.8            & 0.09           & 83.4            & 0.0            & 5.8            & 0.05           & 99.5            & 0.2            & 7.0            & 0.06           & 80.5            & 2.8            & 10.8           & 0.14           & 70.2            \\
                \logo{gemini}Gemini 3.1 Pro~\cite{Gemini_3_1_2026}       & \cmark & 3.4            & 22.9           & 0.18           & 92.4            & 0.0            & \textbf{14.0}  & \ul{0.06}      & 99.5            & 2.5            & 22.5           & 0.18           & 84.5            & 7.8            & 32.2           & 0.32           & 93.2            \\
                \logo{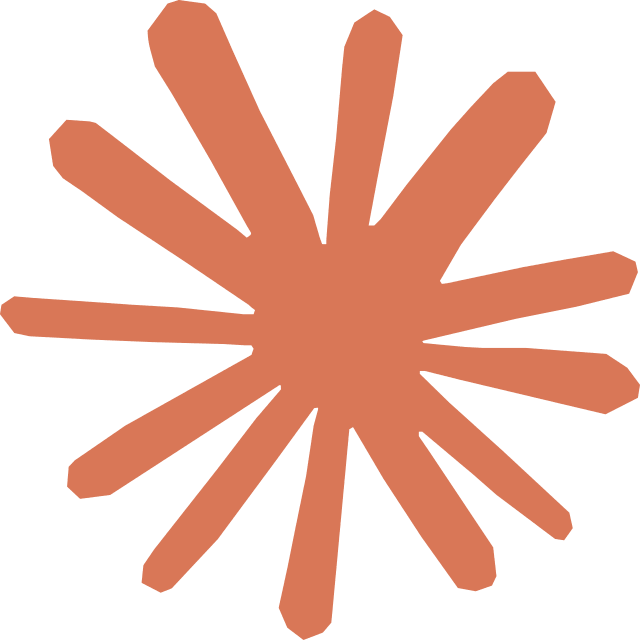}Claude Opus 4.7~\cite{Claude_Opus_4_7_2026} & \cmark & 0.5            & 11.9           & 0.10           & 89.3            & 0.0            & 4.8            & 0.04           & 93.8            & 0.1            & 9.5            & 0.05           & 80.5            & 1.4            & 21.5           & 0.21           & 93.8            \\
                \bottomrule
            \end{tabular}
        }
    }
    \vspace{-10pt}
    \label{tab:archaic_results}
\end{table*}

%% file: tables/mature_results.tex
\begin{table*}[t]
    \centering
    \captionsetup{font={small}}
    \caption{
        \textbf{Evaluation Results on Mature Scripts (Clerical, Regular, Running, Cursive).}
        Performance across two valid tasks: Ancient Text Parsing (Pars.) and Script Classification (Class.). Detection and Spotting are not evaluated for these stages due to the sequence-level annotation paradigm. \textbf{Bold} indicates the best performance while \ul{underlined} results denote the second-best.
    }
    \vspace{-0.8em}
    {
        \setlength{\tabcolsep}{8pt}
        \resizebox{\textwidth}{!}{
            \begin{tabular}{l c | cc | cc | cc | cc | cc}
                \toprule
                \multirow{2}{*}[-1.0mm]
                {\textbf{Model}}                                         &
                \multirow{2}{*}[-1.0mm]
                {\textbf{Think}}                                         &
                \multicolumn{2}{c|}
                {\textbf{Average}}                                       &
                \multicolumn{2}{c|}
                {\textbf{Clerical Script}}                               &
                \multicolumn{2}{c|}
                {\textbf{Regular Script}}                                &
                \multicolumn{2}{c|}
                {\textbf{Running Script}}                                &
                \multicolumn{2}{c}
                {\textbf{Cursive Script}}
                \\
                \cmidrule(lr){3-4}
                \cmidrule(lr){5-6}
                \cmidrule(lr){7-8}
                \cmidrule(lr){9-10}
                \cmidrule(lr){11-12}
                                                                         &        & \textbf{Pars.} & \textbf{Class.} & \textbf{Pars.} & \textbf{Class.} & \textbf{Pars.} & \textbf{Class.} & \textbf{Pars.} & \textbf{Class.} & \textbf{Pars.} & \textbf{Class.} \\
                \midrule
                \multicolumn{12}{c}{\emph{\quad \textbf{Open-Source Models}}}                                                                                                                                                                                    \\
                \logo{intern}InternVL3.5-8B~\cite{InternVL3_5_2025}      &        & 0.40           & 39.8            & 0.41           & 1.8             & 0.51           & 69.4            & 0.38           & 52.9            & 0.30           & 35.0            \\
                \logo{intern}InternVL3.5-A28B~\cite{InternVL3_5_2025}    &        & 0.56           & 63.1            & 0.54           & 28.5            & 0.69           & 85.5            & 0.56           & 63.3            & 0.46           & 75.2            \\
                \logo{qwen}Qwen2.5-VL-7B~\cite{Qwen2_5_VL_2025}          &        & 0.45           & 38.0            & 0.54           & 8.0             & 0.62           & 17.0            & 0.42           & 36.4            & 0.21           & 90.5            \\
                \logo{qwen}Qwen2.5-VL-72B~\cite{Qwen2_5_VL_2025}         &        & 0.49           & 63.0            & 0.59           & 18.0            & 0.66           & 91.5            & 0.46           & 56.6            & 0.26           & 86.0            \\
                \logo{qwen}Qwen3-VL-2B~\cite{Qwen3_VL_2025}              &        & 0.56           & 37.0            & 0.61           & 5.5             & 0.71           & 11.8            & 0.50           & 37.9            & 0.42           & 93.0            \\
                \logo{qwen}Qwen3-VL-8B~\cite{Qwen3_VL_2025}              &        & 0.66           & 67.5            & 0.69           & 32.5            & 0.77           & \textbf{97.2}   & 0.64           & 59.1            & 0.56           & 81.0            \\
                \logo{qwen}Qwen3-VL-8B~\cite{Qwen3_VL_2025}              & \cmark & 0.50           & 50.1            & 0.52           & 11.2            & 0.64           & 79.7            & 0.51           & 53.4            & 0.32           & 56.2            \\
                \logo{qwen}Qwen3-VL-A22B~\cite{Qwen3_VL_2025}            &        & 0.67           & 70.6            & 0.69           & 36.5            & 0.73           & \ul{95.5}       & 0.66           & 68.3            & 0.59           & 82.0            \\
                \logo{qwen}Qwen3-VL-A22B~\cite{Qwen3_VL_2025}            & \cmark & 0.65           & 66.2            & 0.67           & 31.0            & 0.75           & 93.5            & 0.65           & 62.3            & 0.54           & 78.0            \\
                \logo{qwen}Qwen3.5-A3B~\cite{Qwen3_5_2026}               &        & 0.71           & 70.2            & 0.79           & 36.8            & 0.81           & 84.2            & 0.68           & 75.6            & 0.57           & 84.2            \\
                \logo{qwen}Qwen3.5-A17B~\cite{Qwen3_5_2026}              &        & \textbf{0.74}  & 74.5            & \textbf{0.81}  & 52.0            & \ul{0.81}      & 81.3            & 0.67           & 75.3            & \textbf{0.66}  & 89.4            \\
                \logo{gemma}Gemma 4 31B it~\cite{Gemma_2024}             &        & 0.34           & 60.3            & 0.37           & 9.6             & 0.56           & 81.9            & 0.33           & 65.0            & 0.09           & 84.5            \\
                \logo{minicpm}MiniCPM-V 4.5~\cite{MiniCPM_V_4_5_2025}    & \cmark & 0.40           & 49.0            & 0.45           & 2.8             & 0.61           & 87.5            & 0.38           & 56.9            & 0.15           & 48.8            \\
                \logo{AllenAI}Molmo 7B-D 0924~\cite{Molmo_PixMo_2025}    &        & 0.01           & 18.8            & 0.01           & \textbf{70.8}   & 0.01           & 3.0             & 0.01           & 0.7             & 0.01           & 0.5             \\
                \logo{AllenAI}Molmo 72B 0924~\cite{Molmo_PixMo_2025}     &        & 0.00           & 9.8             & 0.00           & 6.8             & 0.01           & 16.5            & 0.01           & 3.2             & 0.00           & 12.8            \\
                \logo{alibaba}Ovis2.6-30B-A3B~\cite{Ovis2_5_2025}        & \cmark & 0.54           & 42.6            & 0.54           & 8.5             & 0.63           & 77.9            & 0.57           & 71.6            & 0.42           & 12.2            \\
                \logo{glmv}GLM-4.5V 108B~\cite{GLM_4_5V_2025}            & \cmark & 0.43           & 60.2            & 0.45           & 11.5            & 0.61           & 84.5            & 0.44           & 63.3            & 0.23           & 81.5            \\
                \logo{kimi}Kimi K2.5~\cite{Kimi_K2_5_2026}               &        & \ul{0.72}      & \textbf{78.1}   & 0.73           & \ul{70.2}       & 0.78           & 78.2            & \ul{0.72}      & \ul{77.8}       & \textbf{0.66}  & 86.0            \\
                \logo{kimi}Kimi K2.5~\cite{Kimi_K2_5_2026}               & \cmark & 0.70           & 75.1            & 0.75           & 68.5            & 0.78           & 81.7            & 0.60           & 65.3            & \textbf{0.66}  & 84.8            \\
                \midrule
                \multicolumn{12}{c}{\emph{\quad \textbf{Proprietary Models}}}                                                                                                                                                                                    \\
                \logo{openai}GPT-4o~\cite{GPT_4o_2023}                   &        & 0.29           & 59.9            & 0.35           & 20.5            & 0.47           & 83.0            & 0.24           & 55.6            & 0.12           & 80.5            \\
                \logo{openai}GPT-5~\cite{GPT_5_2025}                     &        & 0.37           & 61.2            & 0.50           & 36.2            & 0.57           & 59.6            & 0.21           & \textbf{78.1}   & 0.18           & 71.0            \\
                \logo{bytedance}Seed1.8~\cite{Seed_1_8_2026}             &        & 0.69           & 73.1            & 0.68           & 45.5            & 0.79           & 92.7            & 0.69           & 71.8            & 0.61           & 82.5            \\
                \logo{bytedance}Seed1.8~\cite{Seed_1_8_2026}             & \cmark & 0.66           & 72.8            & 0.69           & 48.0            & 0.78           & 89.2            & 0.57           & 73.3            & 0.60           & 80.8            \\
                \logo{bytedance}Seed2.0 Pro~\cite{Seed_2_0_2026}         &        & \ul{0.72}      & \textbf{78.1}   & 0.75           & 60.8            & 0.81           & 82.0            & \textbf{0.73}  & 77.6            & \ul{0.62}      & 92.2            \\
                \logo{bytedance}Seed2.0 Pro~\cite{Seed_2_0_2026}         & \cmark & \ul{0.72}      & \ul{76.8}       & 0.76           & 61.8            & 0.80           & 82.0            & 0.65           & 74.3            & \textbf{0.66}  & 89.0            \\
                \logo{xiaomimimo}MiMo-V2-Omni~\cite{Mimo_V2_Omni_2026}   & \cmark & 0.56           & 64.6            & 0.62           & 40.0            & 0.71           & 80.7            & 0.58           & 73.3            & 0.36           & 64.2            \\
                \logo{gemini}Gemini 2.5 Pro~\cite{Gemini_2_5_2025}       & \cmark & 0.53           & 56.8            & 0.67           & 33.2            & 0.72           & 39.6            & 0.49           & 59.4            & 0.23           & \ul{95.0}       \\
                \logo{gemini}Gemini 3.1 Pro~\cite{Gemini_3_1_2026}       & \cmark & 0.70           & 72.6            & \ul{0.80}      & 61.0            & \textbf{0.83}  & 62.7            & 0.66           & 71.1            & 0.52           & \textbf{95.8}   \\
                \logo{claude}Claude Opus 4.7~\cite{Claude_Opus_4_7_2026} & \cmark & 0.49           & 66.8            & 0.53           & 50.2            & 0.63           & 74.4            & 0.44           & 56.6            & 0.38           & 86.0            \\
                \bottomrule
            \end{tabular}
        }
    }
    \vspace{-10pt}
    \label{tab:mature_results}
\end{table*}

%% file: figure_code/spotting_vis.tex
\begin{figure}[h]
    \centering
    \includegraphics[width=\linewidth]{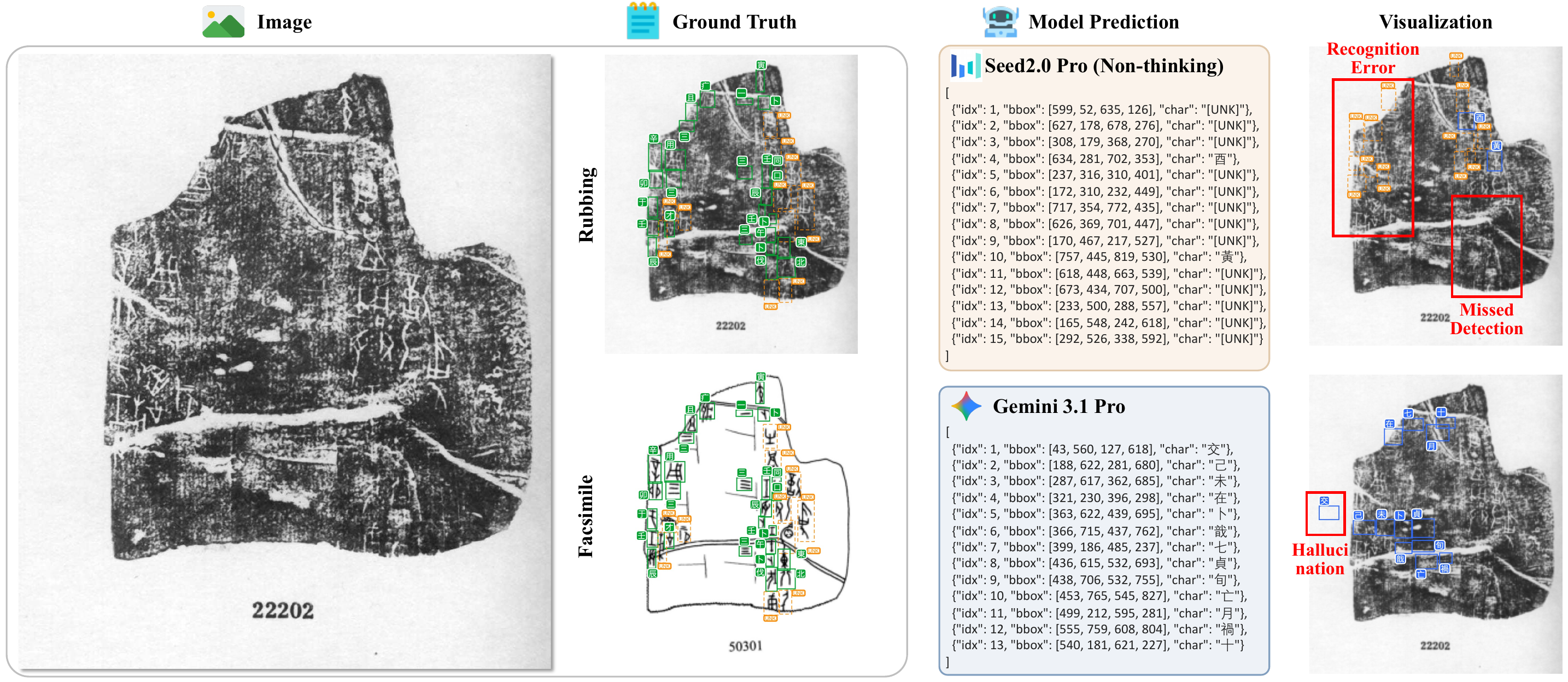}
    \vspace{-1.8em}
    \captionsetup{font={small}}
    \caption{
        \textbf{Qualitative Spotting Results on Oracle Bone Script.}
        Compared to the ground truth, leading VLLMs (Seed2.0 Pro and Gemini 3.1 Pro) struggle with three primary failure modes (highlighted in red): \textit{missed detections} of unconstrained symbols, \textit{recognition errors} due to semantic gaps, and \textit{hallucinations} triggered by physical noise.
    }
    \label{fig:spotting_vis}
\end{figure}

%% file: figure_code/fine_grain_vis.tex
\begin{figure}[h]
    \centering
    \includegraphics[width=\linewidth]{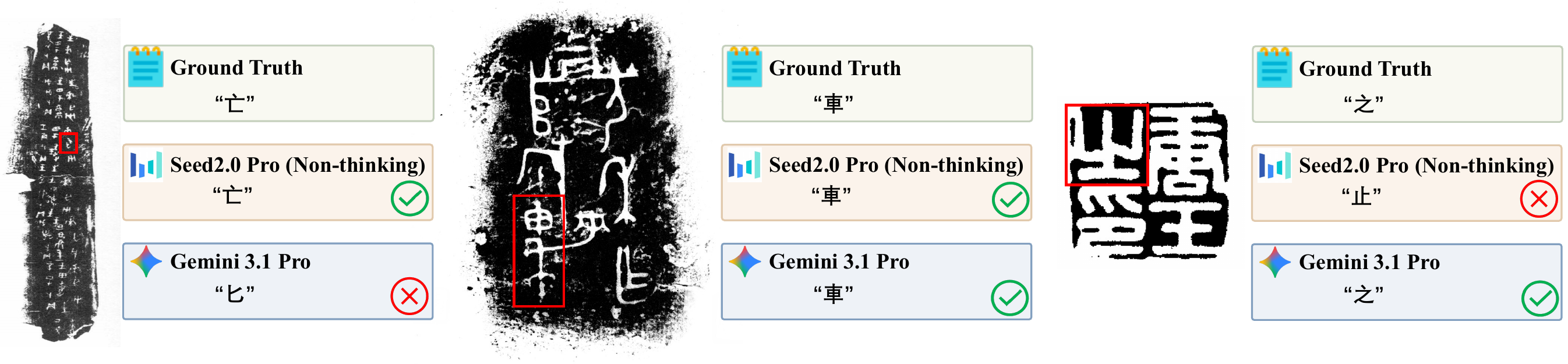}
    \vspace{-1.8em}
    \captionsetup{font={small}}
    \caption{
        \textbf{Qualitative Results of Fine-grained Archaic Character Recognition.}
        By utilizing a visual referring mechanism (\myeg highlighting targets with colored bounding boxes), this task isolates pure morphological decipherment from spatial localization. Despite explicit visual guidance, current VLLMs still struggle to bridge the profound semantic gap between ancient pictographic glyphs and modern characters.
    }
    \label{fig:fine_grain_vis}
    \vspace{-0.8em}
\end{figure}

%% file: figure_code/parsing_vis.tex
\begin{figure}[h]
    \centering
    \includegraphics[width=\linewidth]{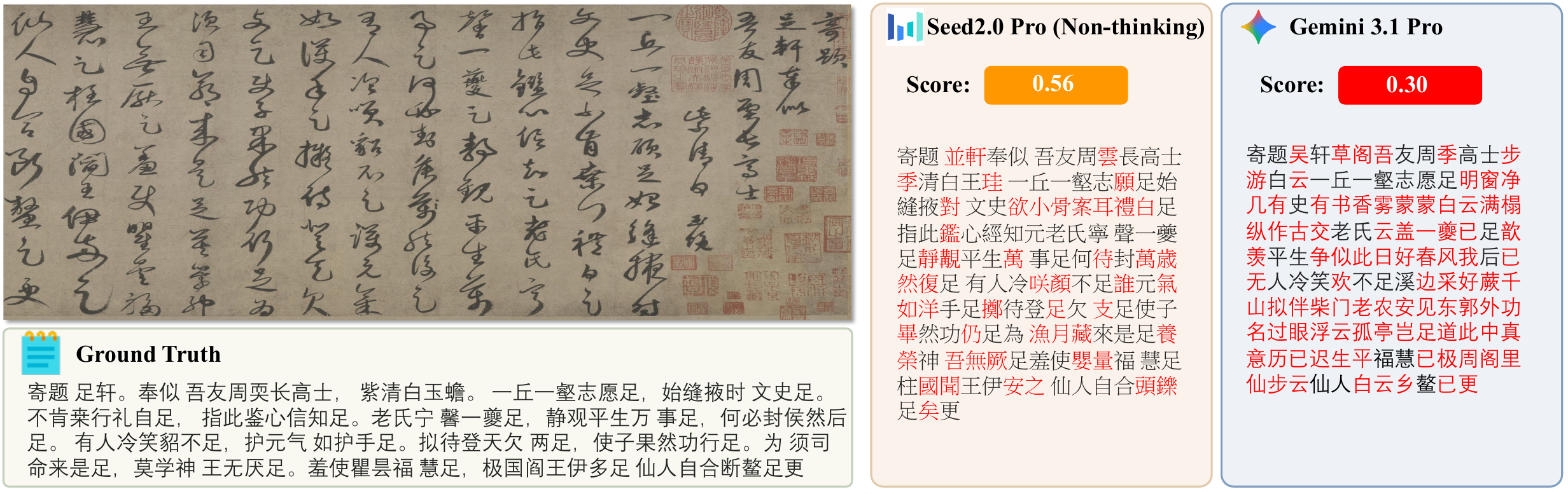}
    \vspace{-1.8em}
    \captionsetup{font={small}}
    \caption{
        \textbf{Qualitative Results of Ancient Text Parsing on Mature Scripts.}
        Even though mature scripts possess standardized layout priors, leading VLLMs still struggle to achieve perfect parsing. As shown in the generated transcriptions (errors highlighted in red), models frequently hallucinate or misinterpret complex continuous strokes, leading to suboptimal NED scores.
    }
    \label{fig:parsing_vis}
\end{figure}

%% file: figure_code/class_vis.tex
\begin{figure}[h]
    \centering
    \includegraphics[width=\linewidth]{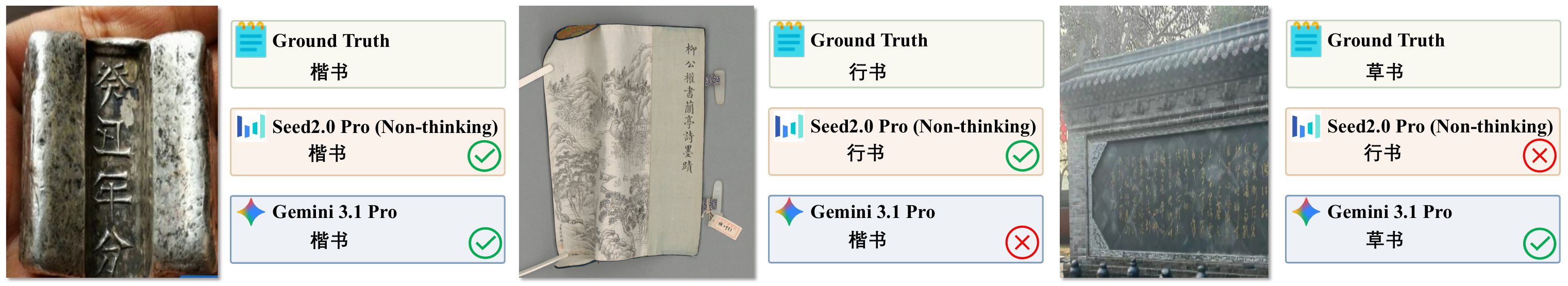}
    \vspace{-1.8em}
    \captionsetup{font={small}}
    \caption{
        \textbf{Visualization of Script Classification Performance.}
        Models reliably categorize archaic scripts by exploiting macro-level textural priors (\myeg shells, bronzes). Conversely, they exhibit severe confusion among mature scripts due to their fundamental inability to differentiate subtle stroke dynamics (\myeg cursiveness) on shared physical mediums.
    }
    \label{fig:class_vis}
\end{figure}

%% file: section/5_conclusion.tex
\section{Conclusion}
\label{sec:conclusion}

In this paper, we introduced \textbf{Chronicles-OCR}, the first comprehensive benchmark designed to evaluate the cross-temporal visual perception capabilities of VLLMs across the evolutionary trajectory of the ``Seven Chinese Scripts.'' Through a novel Stage-Adaptive Annotation Paradigm and four distinct evaluation tasks, including cross-period character spotting, fine-grained archaic character recognition, ancient text parsing, and script classification, we revealed critical bottlenecks in contemporary models when confronting historical texts. Most notably, current VLLMs exhibit catastrophic failures in the fine-grained spatial grounding and semantic decipherment of archaic, unstandardized scripts. Our evaluations further uncovered a profound perception paradox: VLLMs can leverage macro-level stylistic and material priors to accurately classify archaic scripts, yet they severely lack the micro-level stroke perception needed to distinguish mature scripts and struggle to parse unconstrained ancient layouts. By demonstrating that modern document parsing capabilities do not naturally generalize to historically evolved writing systems, Chronicles-OCR provides clear optimization directions and aims to catalyze future research toward robust, evolution-aware multimodal foundation models for Digital Humanities.